\documentclass{article}

    \PassOptionsToPackage{numbers, compress, sort}{natbib}

\usepackage[preprint]{neurips_2022}

\usepackage[utf8]{inputenc} 
\usepackage[T1]{fontenc}    
\usepackage{hyperref}       
\usepackage{url}            
\usepackage{booktabs}       
\usepackage{amsfonts}       
\usepackage{nicefrac}       
\usepackage{microtype}      
\usepackage{xcolor}         

\usepackage{multirow}       
\usepackage{amsmath}        
\usepackage{wrapfig}        
\usepackage{array}
\newcolumntype{P}[1]{>{\centering\arraybackslash}p{#1}}
\newcolumntype{M}[1]{>{\centering\arraybackslash}m{#1}}     
\usepackage{graphbox}       
\usepackage{soul}
\newcommand{\mathcolorbox}[2]{\colorbox{#1}{$\displaystyle #2$}}

\usepackage{graphicx}
\usepackage{bm}
\usepackage{subfig}
\usepackage{bbm}
\usepackage[frozencache,cachedir=.]{minted}

\usepackage{amsmath}
\newcommand{\comb}[2]{{}_{#1}\mathrm{C}_{#2}}

\title{Enhancing Contrastive Learning with \\ Efficient Combinatorial Positive Pairing} 

\author{
  Jaeill~Kim\textsuperscript{1}$^*$, ~Duhun~Hwang\textsuperscript{1}$^*$, ~Eunjung~Lee\textsuperscript{3}$^{\dag}$, \\
  \textbf{~Jangwon~Suh\textsuperscript{1}}, \textbf{~Jimyeong~Kim\textsuperscript{1}}, \textbf{~Wonjong~Rhee\textsuperscript{1,2}}\\
  \textsuperscript{1}Department of Intelligence and Information, Seoul National University,\\
  \textsuperscript{2}Interdisciplinary Program in Artificial Intelligence (IPAI), Seoul National University,\\
  \textsuperscript{3}Artificial Intelligence Cardiovascular Innovation Research, Mayo Clinic\\
  \texttt{jaeill0704@snu.ac.kr}, ~~\texttt{yelobean@snu.ac.kr}, ~~\texttt{lee.eunjung@mayo.edu},\\
  \texttt{rxwe5607@snu.ac.kr}, ~~\texttt{wlaud1001@snu.ac.kr}, ~~\texttt{wrhee@snu.ac.kr}
  }

\begin{document}
\def\thefootnote{*}\footnotetext{Equal contribution.}\def\thefootnote{\arabic{footnote}}
\def\thefootnote{\dag}\footnotetext{Work performed while at Seoul National University.}\def\thefootnote{\arabic{footnote}}

\maketitle

\begin{abstract}
In the past few years, contrastive learning has played a central role for the success of visual unsupervised representation learning. Around the same time, high-performance non-contrastive learning methods have been developed as well. While most of the works utilize only two views, we carefully review the existing multi-view methods and propose a general multi-view strategy that can improve learning speed and performance of any contrastive or non-contrastive method. We first analyze CMC's full-graph paradigm and empirically show that the learning speed of $K$-views can be increased by $\comb{K}{2}$ times for small learning rate and early training. Then, we upgrade CMC's full-graph by mixing views created by a crop-only augmentation, adopting small-size views as in SwAV multi-crop, and modifying the negative sampling. The resulting multi-view strategy is called ECPP~(Efficient Combinatorial Positive Pairing). We investigate the effectiveness of ECPP by applying it to SimCLR and assessing the linear evaluation performance for CIFAR-10 and ImageNet-100. For each benchmark, we achieve a state-of-the-art performance. In case of ImageNet-100, ECPP boosted SimCLR outperforms supervised learning.

\end{abstract}

\section{Introduction}
\label{sec:introduction}

\begin{table}[t!]
\caption{Summary of positive pairing strategies. $K$ views of an image are generated by a stochastic augmentation function. As in the full graph strategy of CMC~\cite{tian2020contrastive}, ECPP utilizes the maximum number of positive pairs, $\comb{K}{2}=\frac{K(K-1)}{2}$. As in the multi-crop of SwAV~\cite{caron2020unsupervised}, ECPP reduces the computational burden by employing small additional views. ECPP adopts three additional improvement techniques that are explained in Section~\ref{sec:ECPP}. Note that the computational burden is proportional to the number and size of the views.}
\label{tab:overview}
\centering
\resizebox{0.98\columnwidth}{!}{
\begin{tabular}{llM{3.3cm}M{3.3cm}M{5cm}}
\toprule
Method     & \multicolumn{1}{c}{Augmentation design} & Loss             & Number of positive pairs              & Computation                \\
\midrule
Baseline (2 views)            &    \includegraphics[align=c,height=2cm]{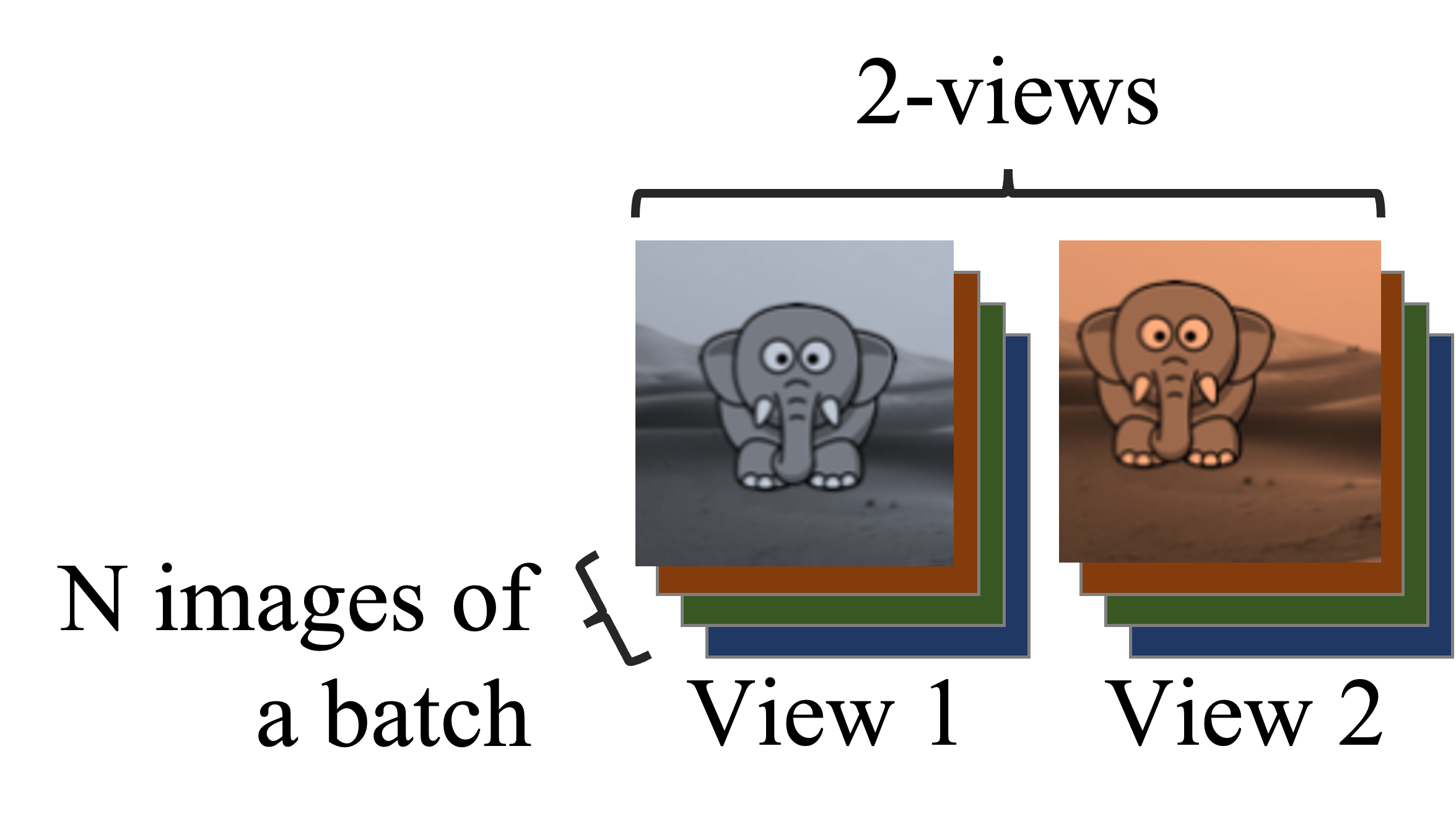}        
& $\mathcal{L}^{V_1,V_2}$ & $N$                            & $2N$                         \\
CMC (core view)   &    \includegraphics[align=c,height=2cm]{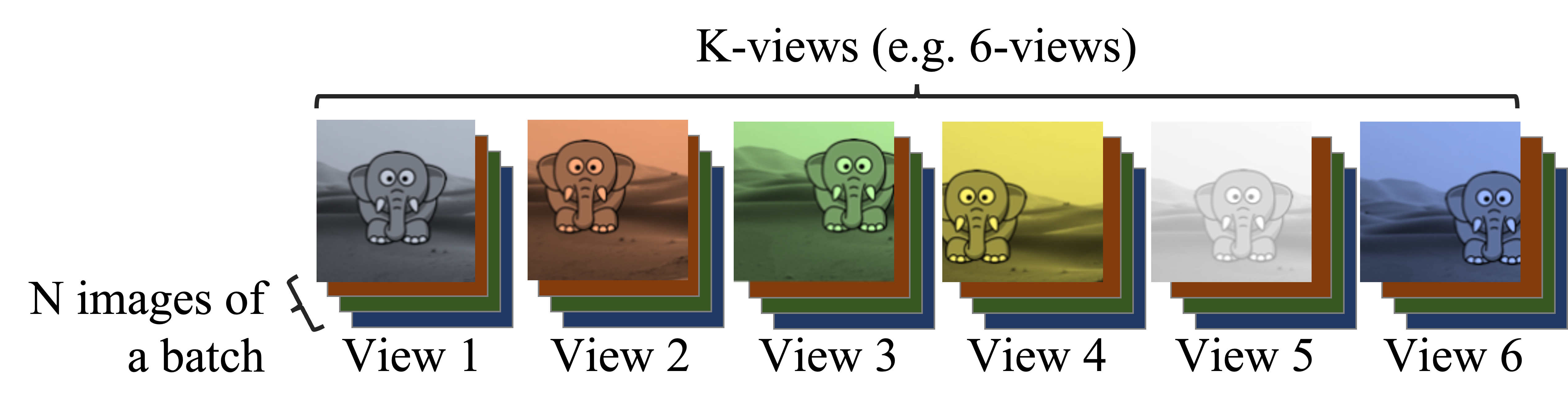}         
& $\sum\limits_{1 < j \le K}\mathcal{L}^{V_1,V_j}$ & $(K-1)N$                       & $K N$                         \\
CMC (full graph)  &    \includegraphics[align=c,height=2cm]{Fig/fullgraphview.png}           
& $\sum\limits_{1\le i <  K}\sum\limits_{i < j \le K}\mathcal{L}^{V_i,V_j}$ & $\mathcolorbox{green}{\frac{K(K-1)}{2}N}$ & $K N$                         \\
Multi-crop              &    \includegraphics[align=c,height=2cm]{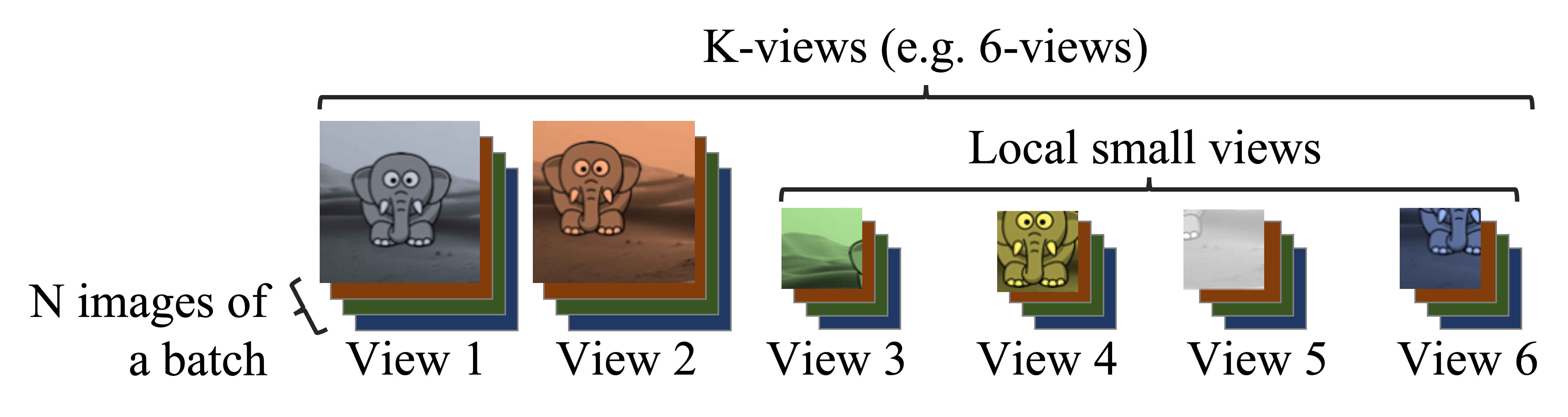}           
& $\sum\limits_{1\le i \le 2}\sum\limits_{i < j \le K}\mathcal{L}^{V_i,V_j}$ & $(2K-3)N$                      & $\mathcolorbox{green}{2N + (K-2)N \cdot \frac{\text{{small image size}}}{\text{{large image size}}}}$ \\
\midrule[0.3pt]\midrule[0.3pt]
ECPP$^{\times K}$ (Ours)   &    \includegraphics[align=c,height=2cm]{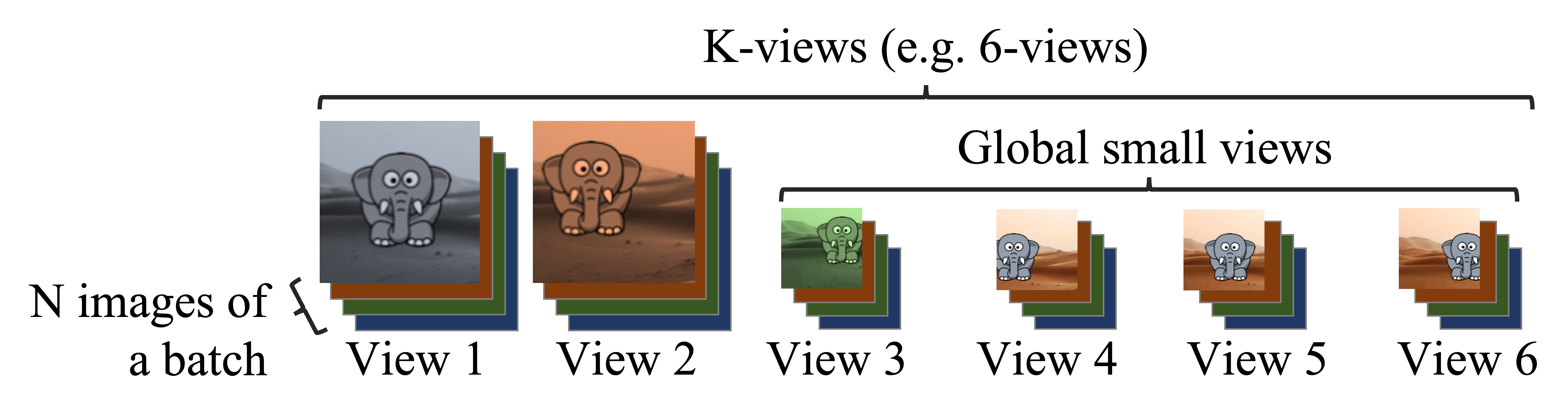}            
& $\sum\limits_{1\le i <  K}\sum\limits_{i < j \le K}\mathcal{L}^{V_i,V_j}$ & $\mathcolorbox{green}{\frac{K(K-1)}{2}N}$ & $\mathcolorbox{green}{2N + (K-2)N \cdot \frac{\text{{small image size}}}{\text{{large image size}}}}$ \\
\bottomrule
\vspace{-1.15cm}
\end{tabular}
}
\end{table}

The early development of visual unsupervised representation learning was typically based on pretext learning techniques applied to a single-view~\cite{doersch2015unsupervised,pathak2016context,noroozi2016unsupervised,gidaris2018unsupervised}. Recent works have achieved an impressive advancement based on contrastive and non-contrastive learning techniques, where typically multi-views are adopted instead of a single-view. Most of the multi-view works, however, are limited in that they focus on two views only. Among the very few works that consider more than two views, CMC~\cite{tian2020contrastive} and SwAV~\cite{caron2020unsupervised} are the most widely known methods. A summary of multi-view methods with more than two views is provided in Table~\ref{tab:overview}.

CMC~\cite{tian2020contrastive} investigates the classic hypothesis that a powerful representation is one that models view-invariant factors, and studies the framework of multi-view contrastive learning.
Because they specifically consider the settings that require more than two views, they have extended the two-view framework to a general multiple-view framework.
For handling more than two views, they suggest \textit{core view} paradigm and \textit{full graph} paradigm. Between the two, full graph paradigm considers combinatorial pairing of $K$ views. 
Following CMC, SimCLR~\cite{chen2020simple} was developed where self-augmentation of an image is used for generating multiple views. 
In SimCLR and most of the following works, use of two views turns out to be simple and powerful as long as a proper augmentation function is chosen. The core view and full graph paradigms of CMC, however, have been mostly forgotten and researchers have been focusing on two-view methods.

SwAV~\cite{caron2020unsupervised} is another important prior-art that utilizes more than two views. SwAV does not require pairwise comparisons, but instead enforces consistency between cluster assignments produced for different augmentation views of the same image. 
They also propose a new data augmentation strategy, \textit{multi-crop}, that uses a mix of views with different resolutions in place of full-resolution views for all. This has the advantage of minimizing the memory and computation overheads that are inevitably created by the additional views.
The main purpose of introducing multi-views in SwAV is to have many views that are known to belong to the same cluster. The design of two full-resolution views with many small additional views allows an increase in the number of views while incurring a relatively small increase in the memory and computation overheads. The two full-resolution views are created with the standard SimCLR augmentation, and the small additional views are created with a slight modification in the crop setting. Multi-crop only considers pairing between one of global views and one of small views, and therefore utilizes less pairs than the combinatorial pairing of CMC's full graph. As can be seen in Table~\ref{tab:overview}, CMC's full graph generates $O(K^2)$ of positive pairs and SwAV's multi-crop generates $O(K)$ of positive pairs. Recently, ReLICv2~\cite{tomasev2022pushing} has adopted multi-crop from SwAV and surpassed the supervised learning performance of the ImageNet-1K classification task.

In this work, we focus on generalizing and improving the previous works on multi-views. 
To be specific, we first focus on understanding the fundamental benefits of $K$-views and then we develop an efficient method called \textit{Efficient Combinatorial Positive Pairing}~(ECPP) for fully and efficiently utilizing $K$-views. ECPP can be combined with any existing contrastive learning or non-contrastive learning algorithm as long as the base algorithm utilizes positive pairing. In our study, we focus on applying ECPP with SimCLR as the base algorithm because of SimCLR's simplicity and broad applicability. As for the notation, we denote SimCLR combined with $K$-view ECPP as SimCLR$^{\times K}$. Similarly, we denote BYOL combined with $K$-view ECPP as BYOL$^{\times K}$. 

We first focus on understanding a fundamental benefit of $K$-views. In CMC~\cite{tian2020contrastive}, it is explained that the enhanced performance, that improves as $K$ increases, can be attributed to the Mutual Information~(MI) maximization between views and to the information minimization that discards nuisance factors that are not shared across the views. The explanations, however, have been challenged in~\cite{tschannen2019mutual,jacobsen2018revnet} for a few reasons including the fact that unsupervised contrastive learning can be successfully performed with an invertible network that has no effect on mutual information. Success of many non-contrastive methods that are not based on InfoNCE (or any other information-theoretic loss) is another reason to suspect the MI based explanations. Instead of the MI explanations, we investigate if the increase in the number of positive pairing terms can provide a simple yet fundamental explanation. Empirically, we show that the performance improvement can be purely dependent on the number of positive pairing terms shown in Table~\ref{tab:overview}. Then, the result has an implication that CMC full graph and ECPP can be exactly 
$\comb{K}{2}=\frac{K(K-1)}{2}$ times faster in learning when compared to a 2-view baseline. 

\setlength\intextsep{0pt}
\begin{wrapfigure}[16]{r}{0.51\linewidth}
\vspace{-0.5cm}
\includegraphics[width=0.49\textwidth]{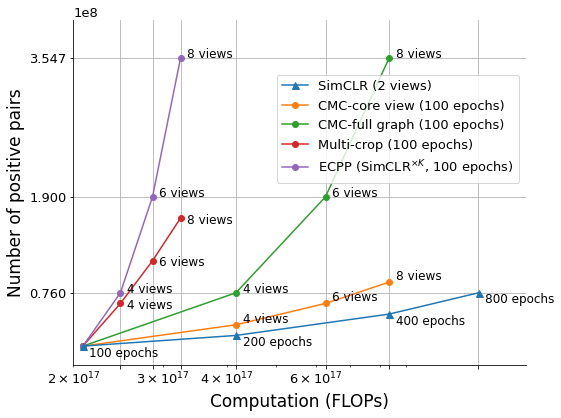}
  \caption{\label{figure:efficiency}
    The number of positive pairs (equivalent to the number of loss terms $\mathcal{L}^{V_i,V_j}$) processed by multi-view representation learning frameworks.
  }
\end{wrapfigure}
\setlength\intextsep{12pt}

With a better understanding on $K$-views, we build up on the previously developed methods of CMC's full graph and SwAV's multi-crop. The design of ECPP starts by combining the $\frac{K(K-1)}{2}$ benefit of CMC with the memory and computation efficiency of multi-crop. As the result, ECPP can process more positive pairings than any other methods for a given amount of computation budget as shown in Figure~\ref{figure:efficiency}. As we will show later, the number of loss terms $\mathcal{L}^{V_i,V_j}$ that are applied during the learning is equal to the number of positive pairs.
To further improve the performance and complete the design of ECPP, we carefully investigate the choice of augmentations for the small additional views, global view vs. local view, and negative sampling scheme.

\vspace{-0.2cm}
\section{Related works}
\label{sec:related_works}
\vspace{-0.2cm}

The recent progress in unsupervised representation learning is primarily driven by improved loss design, efficient optimization, intensive investigations on augmentation design, and selective negative sampling.

\vspace{-0.1cm}

\paragraph{Loss design:} In most cases, a loss is designed to learn invariant representation while preventing representation collapse. To learn invariant representation, most methods maximize the cosine similarity between the two embeddings of a positive pair~\cite{chen2020simple,he2020momentum,chen2021exploring} or cosine similarity's equivalents such as normalized dot product~\cite{grill2020bootstrap}, alignment~\cite{wang2020understanding}, and tolerance~\cite{wang2021understanding}. Barlow Twins~\cite{zbontar2021barlow} maximizes correlation, VICReg~\cite{bardes2021vicreg} maximizes covariance, and SwAV~\cite{caron2020unsupervised} maximizes cross entropies between two positive embeddings. 
A naive implementation of learning invariant representation can end up with a representation collapse. Prevention of representation collapse can be achieved with negative samples~\cite{chen2020simple,he2020momentum}, equal clustering constraint~\cite{caron2020unsupervised}, stop gradient with predictor~\cite{grill2020bootstrap,chen2021exploring}, feature de-correlation~\cite{zbontar2021barlow,bardes2021vicreg}, or uniformity among features on a unit hypersphere~\cite{wang2020understanding}. 

\vspace{-0.1cm}

\paragraph{Optimization improvement:} 
It has been widely observed that the quality of unsupervised representation learning benefits from longer training, large batch sizes, and larger models~\cite{chen2020simple,caron2020unsupervised,grill2020bootstrap,chen2021exploring,he2020momentum,chen2020big,zbontar2021barlow,bardes2021vicreg}. Therefore, a substantial effort has been made to deal with the associated computational challenges.
When training with multiple GPUs via distributed data parallelism~\cite{li2020pytorch}, it has been found that smaller working batch sizes of batch normalization~(BN) can be a problem~\cite{wu2018group} and should be converted to synchronized cross-GPU batch normalization (SyncBN)~\cite{zhang2018context}.
To explicitly reduce the computing FLOPs, mixed precision training~\cite{micikevicius2017mixed} has become a default choice. 
Training with a mix of small and large size images can be helpful for improving the performance~\cite{touvron2019fixing}.
For a faster convergence, a large learning rate with cyclical learning rate scheduling can improve speed and performance of the training~\cite{smith2017cyclical,smith2019super}.
Following such works, use of a large learning rate with cosine learning rate decay~\cite{loshchilov2016sgdr} has been widely adopted for accelerating the rate of convergence. 
Also, linear learning rate scaling rule with a warm-up~\cite{goyal2017accurate} and Layer-wise Adaptive Rate Scaling (LARS)~\cite{you2017large} optimizer have been adopted to resolve the unstable optimization difficulties when training with large learning rates and large batch sizes~\cite{masters2018revisiting}.
In our work, we utilize combinatorial positive pairing of multiple views to improve the computational efficiency of the optimization. 

\vspace{-0.1cm}

\paragraph{Augmentation design:} Most of the recent methods rely on two views generated by a carefully chosen augmentation function. 
Augmentation function design was heavily studied when developing SimCLR~\cite{chen2020simple}, and an excellent stochastic composition was suggested where cropping with resizing, horizontal flipping, color jittering, gray scaling, and Gaussian blurring are utilized. The SimCLR augmentation has become the base setting for many subsequent works~\cite{chen2020improved,chen2021exploring,caron2020unsupervised,grill2020bootstrap,zbontar2021barlow,bardes2021vicreg,tomasev2022pushing}, where slight modifications such as crop size/scale adjustment and inclusion of solarization/saliency-masking~\cite{nguyen2019deepusps} have been considered. 
For more than two views, however, hardly anything has been studied regarding multi-view specific aspects. \cite{caron2020unsupervised} is the only study where the base two views are expanded by small additional views in its multi-crop. The additional views are created by cropping small parts in the anchor image.
In our work, we investigate the augmentation design for the additional views when the first two views are generated with SimCLR augmentation.

\vspace{-0.1cm}

\paragraph{Selective negative sampling:} In contrastive learning, the characteristics of negative examples have been found to be influential to the representation learning.
A hard negative is a negative example whose embedding is similar to the embedding of the anchor example. Using hard negatives for learning was found to be beneficial for deep metric learning~\cite{oh2016deep,harwood2017smart,wu2017sampling,suh2019stochastic,iscen2018mining,zheng2019hardness,xuan2020hard} and also for contrastive learning~\cite{robinson2020contrastive,kalantidis2020hard,wang2021understanding,wu2020conditional}.
A false negative is a negative example whose contextual information match the anchor's contextual information. Using false negatives for learning was found to be harmful and should be removed for contrastive learning~\cite{chuang2020debiased,huynh2022boosting,robinson2020contrastive}. For contrastive learning, however, it is generally impossible to distinguish false negatives from hard negatives in an unsupervised manner~\cite{robinson2020contrastive}.
In our work, we show that the multiple views of an anchor image can be interpreted as false negatives and removing them can be beneficial for multi-view contrastive learning.

\vspace{-0.2cm}
\section{A fundamental benefit of exploiting more than two views}
\label{sec:Delving deep into CMC with SimCLR}
\vspace{-0.2cm}

As explained in Section~\ref{sec:introduction}, ECPP can be combined with any representation learning algorithm that relies on positive pairing through instance augmentation. In this section, we investigate SimCLR as the base algorithm and a vanilla version of SimCLR$^{\times K}$ as its $K$-view extension. For SimCLR$^{\times K}$, we show how the number of positive pairings can affect the performance and speed of learning. We first define the loss functions.

\vspace{-0.1cm}

\subsection{Loss of SimCLR ($\mathcal{L}_\text{SimCLR}$)}

\vspace{-0.1cm}

Given an encoding network $f(\cdot)$, a projection network $g(\cdot)$, stochastic augmentation functions $t\sim\mathcal{T}$ and $t'\sim\mathcal{T'}$, and a mini-batch of $N$ examples, SimCLR generates $N$ pairs of embedding vectors. For an example $x_n$, the embedding vectors are $z_n(=g\circ f\circ t(x_n))$ and $z_n'(=g\circ f\circ t'(x_n))$. Therefore, we end up with $2N$ embeddings for the mini-batch. The loss function of SimCLR for an ordered positive pair $(z_n,z_n')$ can be described as
\begin{align}
\vspace{-0.1cm}
\label{eq:simclr_with_false_negative}
\ell_{(z_n,z_n')} &= -log\frac{\text{exp}(\text{sim}(z_n,z_n')/\tau)}{\sum_{m=1}^{N} \mathbbm{1}_{[m\neq n]} \text{exp}(\text{sim}(z_n,z_m)/\tau)+\sum_{m=1}^{N}\text{exp}(\text{sim}(z_n,z_m')/\tau)},
\end{align}
where $\tau$ denotes the temperature parameter and $\text{sim}(z_n,z_n')$ is the cosine similarity that is equivalent to the normalized dot product. Note that $\ell_{(z_n,z_n')} \neq \ell_{(z_n',z_n)}$. Finally, the losses of all $2N$ embeddings are summed to form the SimCLR loss as below. 
\begin{align}
\vspace{-0.1cm}
\label{eq:simclr}
\mathcal{L}_\text{SimCLR}&=\sum_{n=1}^N\ell_{(z_n,z_n')}+\ell_{(z_n',z_n)}
\end{align}

\vspace{-0.3cm}

\subsection{Loss of SimCLR with $K$-views (vanilla version; $\mathcal{L}_{\text{SimCLR}^{\times K}}^{\text{Vanilla}}$)}
SimCLR is based on two views that are generated by a stochastic augmentation function. We can name the two views as $V_1$ and $V_2$. To reflect the views in the notation, we can slightly modify the notation $\mathcal{L}_{\text{SimCLR}}$ in Eq.~\eqref{eq:simclr} into $\mathcal{L}_{\text{SimCLR}}^{V_1,V_2}$. With this updated notation, the loss of SimCLR with $K$-views can be defined as below. 
\begin{align}
\vspace{-0.1cm}
\label{eq:simclr_k_views}
\mathcal{L}_{\text{SimCLR}^{\times K}}^{\text{Vanilla}}&=\sum_{1\le i < j \le K}\mathcal{L}_{\text{SimCLR}}^{V_i,V_j}
\vspace{-0.1cm}
\end{align}
Strictly speaking, the loss in Eq.~\eqref{eq:simclr_k_views} is simply $\mathcal{L}_\text{SimCLR}$ extended with $K$-views and CMC's full graph paradigm. Therefore, it is different from the full ECPP implementation ($\mathcal{L}_{\text{SimCLR}^{\times K}}$) that requires additional modifications that will be explained in Section~\ref{sec:ECPP}. Therefore, we denote the loss as 
$\mathcal{L}_{\text{SimCLR}^{\times K}}^{\text{Vanilla}}$ instead of $\mathcal{L}_{\text{SimCLR}^{\times K}}$. In this section, we want to focus on the effect of combinatorial positive pairing alone without being affected by the other modifications of ECPP, and thus investigate $\mathcal{L}_{\text{SimCLR}^{\times K}}^{\text{Vanilla}}$ instead of $\mathcal{L}_{\text{SimCLR}^{\times K}}$. With this caution, we provide the results of empirical investigations in the following Section~\ref{subsec:learing_speed}. 

\vspace{-0.2cm}
\subsection{Analysis on the speed and performance of learning}
\label{subsec:learing_speed}
\vspace{-0.2cm}
For convolutional neural networks, training with batch size of $N$ for $k$ steps has been shown to perform about the same as the training with batch size of $N\cdot k$ for one step~\cite{krizhevsky2014one, you2017large}. Here, we investigate if a similar result can be obtained for the $K$-view loss in Eq.~\eqref{eq:simclr_k_views}. As summarized in Table~\ref{tab:overview} and as can be checked by comparing Eq.~\eqref{eq:simclr} and Eq.~\eqref{eq:simclr_k_views}, learning with $K$-views can be considered as learning with $\comb{K}{2}$ times more examples. This can be verified by confirming that the loss in Eq.~\eqref{eq:simclr_k_views} is formed by summing $\comb{K}{2}$ of $\mathcal{L}_{\text{SimCLR}}^{V_i,V_j}$ terms. Note that $\comb{K}{2}$ corresponds to the number of positive paring that is possible with $K$ views.

\begin{figure}[tb!]
  \centering
  \subfloat[Epoch \label{figure:efficiency_curve_a}]{
    \includegraphics[width=0.24\textwidth]{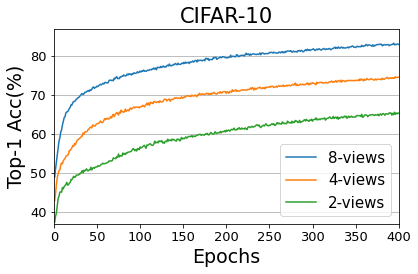}
    \includegraphics[width=0.24\textwidth]{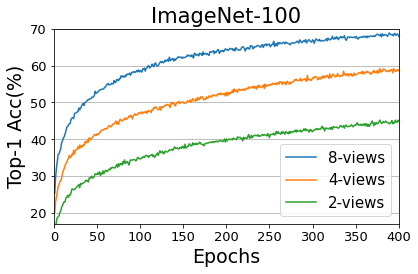}
  }
  \subfloat[Number of positive pairs \label{figure:efficiency_curve_b}]{
    \includegraphics[width=0.24\textwidth]{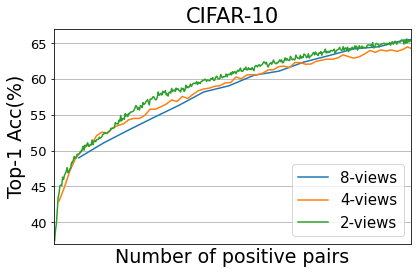}
    \includegraphics[width=0.24\textwidth]{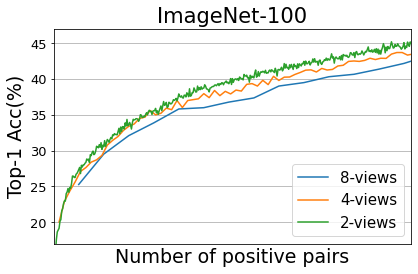}
  }
\caption{\label{figure:efficiency_curve}
Linear evaluation performance of ResNet-18 with 2, 4, and 8 views. (a) SimCLR with more views learns faster. In fact, SimCLR with $K$-views learns $\comb{K}{2}$ times faster for any given iteration number. (b) By adjusting the $X$-axis to the number of processed positive pairs (i.e., the number of $\mathcal{L}_{\text{SimCLR}}^{V_i,V_j}$ terms used for back-propagation), it can be seen that the learning speed is about the same as long as the processed number of positive pairs is the same.}
\vspace{-0.5cm}
\end{figure}

To verify if the learning speed improvement is indeed $\comb{K}{2}$, we performed a contrastive learning using the loss in Eq.~\eqref{eq:simclr_k_views} and the results are shown in Figure~\ref{figure:efficiency_curve}. At the end of each epoch of contrastive learning, we have performed a linear evaluation to assess the quality of the learned representations. To decouple other learning effects, we used a constant learning rate of $0.0004$ without any learning rate scheduling such as warm-up and learning rate decay. We have also used the full-resolution with SimCLR augmentations for all $K$ views. Therefore, the learning speed improvement is due to the use of combinatorial positive pairing alone. From Figure\mbox{~\ref{figure:efficiency_curve}}(a), the performance improvement for the same epoch can be confirmed. From Figure\mbox{~\ref{figure:efficiency_curve}}(b), the speed improvement of $\comb{K}{2}$ can be confirmed. Surprisingly, the learning speed increase of exploiting $K$ views is quite large. The speed increase comes at the cost of processing more views. The computational burden increases linearly with $K$ while the speed benefit increases with $\comb{K}{2}$.

The full ECPP implementation utilizes additional techniques such as the standard acceleration with large learning rate with learning rate scheduling, small additional views, and a loss modification to handle negative pairing. With the additional techniques, the learning dynamics become sophisticated and we will provide a further analysis in Section~\ref{sec:discussion}.

\vspace{-0.2cm}
\section{Efficient combinatorial positive pairing (ECPP)}
\label{sec:ECPP}
\vspace{-0.2cm}

\subsection{Designing augmentation of the additional views}
\label{subsec:augmanetation_design}
\vspace{-0.1cm}

\begin{table}[t!]
\caption{Effectiveness of each augmentation technique. For the baseline 2-view SimCLR on ResNet-50, we measure linear evaluation performance (top-1\% validation accuracy). The first set in the table shows how much change is observed by excluding a technique from the default augmentation of SimCLR. The second set in the table shows the same for including only a single augmentation but nothing else. Crop is most sensitive in both sets, indicating that it is the most influential.}
\label{tab:crop_similarity}
        \centering
        \resizebox{0.42\textwidth}{!}{
        \begin{tabular}{lrr}
            \toprule
            Augmentation        & \multicolumn{1}{l}{CIFAR-10} & \multicolumn{1}{l}{ImageNet-100} \\
            \midrule
            Baseline augmentation & 89.88                        & 78.66                            \\
            \cmidrule(r){1-1} \cmidrule(r){2-3}
            (except) Crop              & \textbf{67.10}                & \textbf{40.56}                   \\
            (except) Flip              & 89.64                        & 76.94                            \\
            (except) ColorJitter       & 85.14                        & 74.40                             \\
            (except) GrayScale         & 86.90                         & 77.20                             \\
            (except) GaussianBlur      & -                            & 78.82                            \\
            (except) Solarization      & -                            & 78.20                             \\
            \midrule
            No augmentation     & \multicolumn{1}{l}{}         & \multicolumn{1}{l}{}             \\
            \cmidrule(r){1-1} \cmidrule(r){2-3}
            (include) Crop              & \textbf{74.23}               & \textbf{50.10}                    \\
            (include) Flip              & 25.22                        & 5.76                                \\
            (include) ColorJitter       & 55.99                        & 23.98                             \\
            (include) GrayScale         & 22.86                        & 4.70                                \\
            (include) GaussianBlur      & -                            & 6.48                                \\
            (include) Solarization      & -                            & 7.92                             \\
            \bottomrule
        \end{tabular}
        }
        \vspace{-0.2cm}
\end{table}

\cite{chen2020simple, misra2020self} noted that crop plays a crucial role when designing augmentations. To confirm the previous findings and quantitatively investigate the importance of each individual augmentation, we performed a new ablation study for CIFAR-10 and ImageNet-100. The results are shown in Table~\ref{tab:crop_similarity} where crop has the largest impact on the linear evaluation performance.

For $K$-view ECPP, the original two views are kept intact by applying the original augmentation scheme. For the additional views, however, there is no clear reason to apply the same augmentation scheme. It might be better to apply heterogeneous augmentation schemes to increase the diversity in learning. Motivated by the superiority of crop, we investigate mixing SimCLR augmentation with a new augmentation scheme named \textit{crop-only} augmentation. The results are shown in~Table~\ref{tab:croponlyaug}. Based on the results, we design ECPP to apply SimCLR augmentation to the half of the views and crop-only augmentation to the other half.

\begin{table}[t!]
\centering
\caption{The effect of mixed views created by applying the standard SimCLR augmentation or crop-only augmentation. Top-1\% linear evaluation results for ResNet-50 and 100 epochs are shown. Mixing views helps when compared to using SimCLR views only.}
\label{tab:croponlyaug}
\resizebox{0.75\textwidth}{!}{
\begin{tabular}{@{}cccccc@{}}
\toprule
            & \multicolumn{2}{c}{Augmentation design} &       \multicolumn{2}{c}{Dataset} \\ 
\midrule
Total number of views & Number of SimCLR views  & Number of crop-only views &  \multicolumn{1}{c}{CIFAR-10}                  &    \multicolumn{1}{c}{ImageNet-100}                  \\ 
\midrule
\multirow{3}{*}{4}  & 4             & 0               & 91.94         & 83.14                     \\
                    & 3             & 1               & 92.27           & 83.50                     \\
                    & 2             & 2               & \textbf{93.07}   & \textbf{84.46}   \\ 
\cmidrule(r){1-1} \cmidrule(r){2-3} \cmidrule(r){4-5} 
\multirow{5}{*}{6}  & 6             & 0               & 92.40            & 84.30                     \\
                    & 5             & 1               & 93.02              & 84.68                     \\
                    & 4             & 2               & 93.05             & 83.90                     \\
                    & 3             & 3               & \textbf{93.50}  & 84.30            \\
                    & 2             & 4               & 93.39              & \textbf{85.10}           \\ 
\cmidrule(r){1-1} \cmidrule(r){2-3} \cmidrule(r){4-5} 
\multirow{7}{*}{8}  & 8             & 0               & 93.00              & 84.18                     \\
                    & 7             & 1               & 93.13                & 84.04                     \\
                    & 6             & 2               & 93.55                  & 84.24                     \\
                    & 5             & 3               & 93.29           & \textbf{84.90}            \\
                    & 4             & 4               & 93.55              & 84.62                     \\
                    & 3             & 5               & \textbf{93.72}    & 84.56            \\
                    & 2             & 6               & 93.17             & 83.86                     \\ 
\bottomrule
\end{tabular}
}
\vspace{-0.3cm}
\end{table}

\subsection{Applying small views of multi-crop to the additional views for efficiency}
\label{subsec:small_views}

The downside of using $K$-views is in the computational side. By creating additional views, the amount of forward computation is increased from $2N$ for the base algorithm (e.g., SimCLR) to $KN$ for the ECPP combined algorithm (e.g., SimCLR$^{\times K}$). The increase in computational burden is by a factor of $\frac{KN}{2N}=\frac{K}{2}$. Multi-crop in~SwAV~\cite{caron2020unsupervised} wisely mitigates this burden by choosing a smaller image size for the additional views while keeping the size of the original two views intact. By reducing each of height and width of an image by a factor of $r$, the reduction in image's area becomes a factor of $r^2$.  

\begin{table}[b!]
\vspace{-0.3cm}
\centering
\caption{\label{tab:crop_scale}The effect of controlling the size of $K-2$ additional views and the hyperparameters of the $\frac{K}{2}$ crop views. For ECPP, we choose small views not only for the computational efficiency but also for a better performance. Also, we choose the crop scale of [0.20, 1.00] (global views are created) instead of the multi-crop's [0.05, 0.14] (local views are created). Top-1\% linear evaluation results for ResNet-50 on ImageNet-100 with 100 epochs are shown.}
\resizebox{0.75\textwidth}{!}{
\begin{tabular}{lclcrrr}
    \toprule
        \multicolumn{3}{c}{Crop design and size of the additional views} & Pairing strategy & 4-views & 6-views & 8-views \\
        \midrule
        Default large view  & {[}0.20, 1.00{]}  &   (224 $\times$ 224)       & $_KC_2$ (baseline) & 83.56   & 83.88 & 83.86  \\
        \cmidrule(r){1-3} \cmidrule(r){4-4} \cmidrule(r){5-7} 
        \multirow{2}{*}{Local small view}  & \multirow{2}{*}{{[}0.05, 0.14{]}} & \multirow{2}{*}{(96 $\times$ 96)}  & $2K-3$ (multi-crop) & 82.28   & 84.00 & 84.26   \\
                       &      &                   & $_KC_2$ & 81.16   & 82.52 & 82.44 \\
        \cmidrule(r){1-3} \cmidrule(r){4-4} \cmidrule(r){5-7}
        \multirow{2}{*}{Global small view} & \multirow{2}{*}{{[}0.20, 1.00{]}} & \multirow{2}{*}{(96 $\times$ 96)}  & $2K-3$ & 83.02   & 84.06 & 84.44  \\
                      & &      & $_KC_2$ (ECPP) & \textbf{84.12}   & \textbf{84.30} & \textbf{84.58} \\
    \bottomrule
\vspace{-0.7cm}
\end{tabular}
}
\end{table}

We explored the effect of image size and the hyperparameters of crop, and the results are shown in Table~\ref{tab:crop_scale}. As for the size, we choose small views as in the multi-crop. Then, the computational burden is reduced from $KN$ to $2N + (K-2)N \cdot \frac{\text{{small image size}}}{\text{{large image size}}}$. For ImageNet, $\frac{\text{{small image size}}}{\text{{large image size}}} = \frac{96^2}{224^2} = 0.18$ which is $82\%$ reduction in the overhead that is created by the additional views. In fact, the use of small views allows a better performance than the baseline. Related to this, it has been already reported that training with a mix of small size images and full size images in multi-crop can improve the classification performance in~\cite{touvron2019fixing}.

As for the hyperparameters of crop, we depart from multi-crop's local view design and use global views that cover almost the entire image using a weak crop scale and a strong resize. Using local views caused performance degradation in our study, especially when combinatorial positive pairing ($_KC_2$) is used. This is most likely due to the reduced probability of the views sharing the same contextual information.

\subsection{A slight modification on negative sampling}
\label{section:false_negative}
The loss of ECPP is almost the same as the loss in Eq.~\eqref{eq:simclr_k_views}, except for one modification. For the loss term $\ell_{(z_n,z_n')}$ in Eq.~\eqref{eq:simclr_with_false_negative}, $z_n$ serves as the anchor. While $z_n$ is not included as a negative sample in the denominator, $z_n'$ is kept according to Eq.~\eqref{eq:simclr_with_false_negative}. However, $z_n'$ is a self-augmented version of $z_n$, and we find it helpful to remove it from the set of negative samples as shown below. 
\begin{align}
\label{eq:simclr_without_false_negative}
\ell'_{(z_n,z_n')} &= -log\frac{\text{exp}(\text{sim}(z_n,z_n')/\tau)}{\sum_{m=1}^{N}\mathbbm{1}_{[m\neq n]}(\text{exp}(\text{sim}(z_n,z_m)/\tau) + \text{exp}(\text{sim}(z_n,z_m')/\tau))}
\end{align}
In conclusion, our final design of ECPP loss, $\mathcal{L}_{\text{SimCLR}^{\times K}}$, is the same as $\mathcal{L}_{\text{SimCLR}^{\times K}}^{\text{Vanilla}}$ except that $\ell_{(z_n,z_n')}$ is replaced with $\ell'_{(z_n,z_n')}$. Performance comparison results are shown in Table~\ref{tab:neg} where the use of $\ell'_{(z_n,z_n')}$ always performs better.

\begin{table}[ht!]
\centering
\vspace{-0.3cm}
\caption{\label{tab:neg}
Comparison between the loss $\mathcal{L}_{\text{SimCLR}^{\times K}}^{\text{Vanilla}}$ in Eq.~\eqref{eq:simclr_k_views} and ECPP's loss $\mathcal{L}_{\text{SimCLR}^{\times K}}$. ECPP uses a slightly difference loss where $\ell'_{(z_n,z_n')}$ in Eq.~\eqref{eq:simclr_without_false_negative} is used instead of $\ell_{(z_n,z_n')}$ in Eq.~\eqref{eq:simclr_with_false_negative}. Top-1\% linear evaluation performance is shown for networks trained for 100 epochs.}
\resizebox{0.7\textwidth}{!}{
\begin{tabular}{@{}lcccccccc@{}}
\toprule
         & \multicolumn{4}{c}{CIFAR-10}                                      & \multicolumn{4}{c}{ImageNet-100}                                  \\ \midrule
         & \multicolumn{2}{c}{ResNet-18}   & \multicolumn{2}{c}{ResNet-50}   & \multicolumn{2}{c}{ResNet-18}   & \multicolumn{2}{c}{ResNet-50}   \\ \midrule
Loss     & 2-views              & 4-views              & 2-views              & 4-views              & 2-views              & 4-views              & 2-views              & 4-views              \\ \midrule
$\ell$ (Eq.\eqref{eq:simclr_with_false_negative}) & 85.87          & 89.43          & 86.31          & 91.88          & 73.44          & 76.62          & 77.28          & 82.68          \\
$\ell'$ (Eq.\eqref{eq:simclr_without_false_negative}) & \textbf{86.12} & \textbf{89.78} & \textbf{87.20} & \textbf{91.94} & \textbf{73.78} & \textbf{77.18} & \textbf{78.30} & \textbf{83.56} \\ \bottomrule
\vspace{-0.7cm}
\end{tabular}

}
\end{table}

\vspace{-0.1cm}
\section{Experimental results}
\label{sec:Experiment}
\vspace{-0.2cm}

We pre-train multi-view representation learning models without labels by implementing the same architecture and training protocols as in SimCLR~\cite{chen2020simple} with three datasets, CIFAR-10~\citep{krizhevsky2009learning}, ImageNet-100 (the subset of randomly chosen 100 classes of ImageNet-1K as in~\cite{tian2020contrastive}), and ImageNet-1K~\citep{russakovsky2015imagenet}. The default baseline augmentation sets for CIFAR-10 and ImageNet-1K (and ImageNet-100) follow the standard SimCLR augmentation sets for CIFAR-10 and ImageNet-1K, respectively. For ImageNet-1K (and ImageNet-100), solarization is applied as the final step. Unless otherwise stated, the default batch size is 64 and the base encoding network is the standard ResNet-50~\citep{he2016deep} where the representation is the output of the penultimate layer of 2048 dimensions. When training with CIFAR-10, we make the standard modifications to the encoding network for tiny images where the first convolution layer is replaced with 3x3 Conv of stride 1 and the first maxpool layer is removed.
When evaluating the models, we follow the standard linear evaluation practice~\cite{alain2016understanding, misra2020self,chen2020simple, grill2020bootstrap, caron2020unsupervised} where the linear classifier on top of the frozen pre-trained model is trained with training dataset in a supervised manner and the classification performance is evaluated with the validation dataset. A complete set of implementation details are provided in Appendix~\ref{section:implementation_details}.

\begin{table}[t]
\caption{\label{tab:rslt1}Comparison between the best known prior arts and SimCLR combined with $K$-view ECPP (i.e.,  SimCLR$^{\times K}$). Top-1\% linear evaluation of classification is shown. All of our ECPP results are obtained with a single hyperparameter setting (see Appendix~\ref{section:implementation_details}). For ImageNet-1K, a proper tuning  of hyperparameters will very likely improve the performance of SimCLR$^{\times K}$. For ImageNet-100, SimCLR$^{\times \text{4}}$ outperforms the supervised learning performance.}
\subfloat[CIFAR-10]{
\label{tab:rslt1_a}
    \resizebox{.32\textwidth}{!}{
\begin{tabular}{lrr}
\toprule
Method & \multicolumn{1}{c}{Epoch} & \multicolumn{1}{c}{Top-1} \\
\midrule
Supervised                     & \multicolumn{1}{l}{}      & 95.0                              \\
\midrule
Decoupled NT-Xent \cite{chen2021intriguing}             & 800                       & 94.0                              \\
SWD \cite{chen2021intriguing}                           & 800                       & 94.1                            \\
\midrule
SimCLR$^{\times 2}$                  & 800                       & 93.9                            \\
SimCLR$^{\times 4}$                  & 800                       & \textbf{94.4}                           \\
SimCLR$^{\times 6}$                  & 800                       & 94.3                            \\
SimCLR$^{\times 8}$                  & 200                       & \textbf{94.4}                           \\
\bottomrule
\vspace{-1.52cm}
\end{tabular}
}

}
\hfill
\subfloat[ImageNet-100]{
\label{tab:rslt1_b}
    \resizebox{.32\textwidth}{!}{
\begin{tabular}{lrr}
\toprule
Method & \multicolumn{1}{c}{Epoch} & \multicolumn{1}{c}{Top-1} \\
\midrule
Supervised                     & \multicolumn{1}{l}{}      & 86.2                            \\
\midrule
Align Uniform \cite{wang2020understanding}                 & 240                       & 74.6                            \\
CMC (K=1) \cite{zheng2021contrastive}                       & 200                       & 75.8                            \\
CMC (K=4) \cite{zheng2021contrastive}                       & 200                       & 78.8                           \\
CACR(K=1) \cite{zheng2021contrastive}                       & 200                       & 79.4                            \\
CACR(K=4) \cite{zheng2021contrastive}                       & 200                       & 80.5                           \\
LooC++ \cite{xiao2020should}                        & 500                       & 82.2                            \\
MoCo-v2+MoCHi \cite{kalantidis2020hard}                 & 800                       & 84.5                            \\
\midrule
SimCLR$^{\times 2}$                  & 800                       & 84.5                           \\
SimCLR$^{\times 4}$                  & 800                       & \textbf{87.0}                           \\
SimCLR$^{\times 6}$                  & 400                       & 86.6                           \\
SimCLR$^{\times 8}$                  & 100                       & 84.6                           \\
SimCLR$^{\times 8}$                  & 200                       & 85.6                            \\
\bottomrule
\vspace{-1.52cm}
\end{tabular}
}

}
\hfill
\subfloat[ImageNet-1K]{
\label{tab:rslt1_c}
    \resizebox{.27\textwidth}{!}{
\begin{tabular}{lrr}
\toprule
Method & \multicolumn{1}{l}{Epoch} & \multicolumn{1}{l}{Top-1} \\
\midrule
Supervised && 76.5\\
\midrule
InstDisc~\cite{hua2021feature}                       & 200                            & 58.5                      \\
LocalAgg~\cite{hua2021feature}                       & 200                            & 58.8                      \\
MoCo~\cite{hua2021feature}                           & 200                            & 60.8                      \\
CPC v2~\cite{hua2021feature}                         & 200                            & 63.8                      \\
Shuffled-DBN~\cite{hua2021feature}                   & 200                            & 65.2                     \\
MoCo v2~\cite{hua2021feature}                        & 200                            & 67.5                      \\
PCL v2~\cite{hua2021feature}                         & 200                            & 67.6                      \\
PIC~\cite{hua2021feature}                            & 200                            & 67.6                      \\
MoCHi~\cite{hua2021feature}                          & 200                            & 68.0                        \\
AdCo~\cite{hua2021feature}                           & 200                            & 68.6                      \\
SwAV~\cite{hua2021feature}                           & 200                            & \textbf{72.7}                      \\
\midrule
SimCLR~\cite{chen2020simple}      & 200                            & 64.3                      \\
SimCLR$^{\times 6}$            & 200                            & 71.1                      \\
\bottomrule
\vspace{-1.52cm}
\end{tabular}
}

}
\end{table}

\vspace{-0.1cm}
\subsection{Comparison with prior arts}
\label{subsec:experiment_comparison}
\vspace{-0.1cm}
Table~\ref{tab:rslt1} shows the comparison results of the linear evaluation. SimCLR$^{\times K}$ not only outperforms the other methods for CIFAR-10 and ImageNet-100 but also surpasses the supervised baseline for ImageNet-100. For ImageNet-1K, SimCLR$^{\times K}$ improves the performance of SimCLR baseline by $6.8\%$. 
All of our results in the table, for all three datasets, were obtained with a single hyperparameter setting (see Appendix~\ref{section:implementation_details}). For ImageNet-1K, it should be possible to improve the performance of SimCLR$^{\times K}$ with an adequate amount of tuning. 

\subsection{The effect of training length (maximum epoch value)}

\begin{figure}[h!]
\vspace{-0.4cm}
\parbox{.5\linewidth}{
\centering
  \subfloat[Maximum epoch configuration]{
    \includegraphics[width=0.38\textwidth]{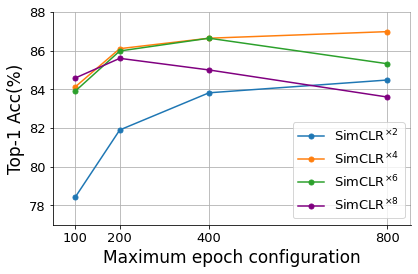}
  }
  }
\parbox{.5\linewidth}{
\centering
  \subfloat[Number of positive pairs]{
    \includegraphics[width=0.38\textwidth]{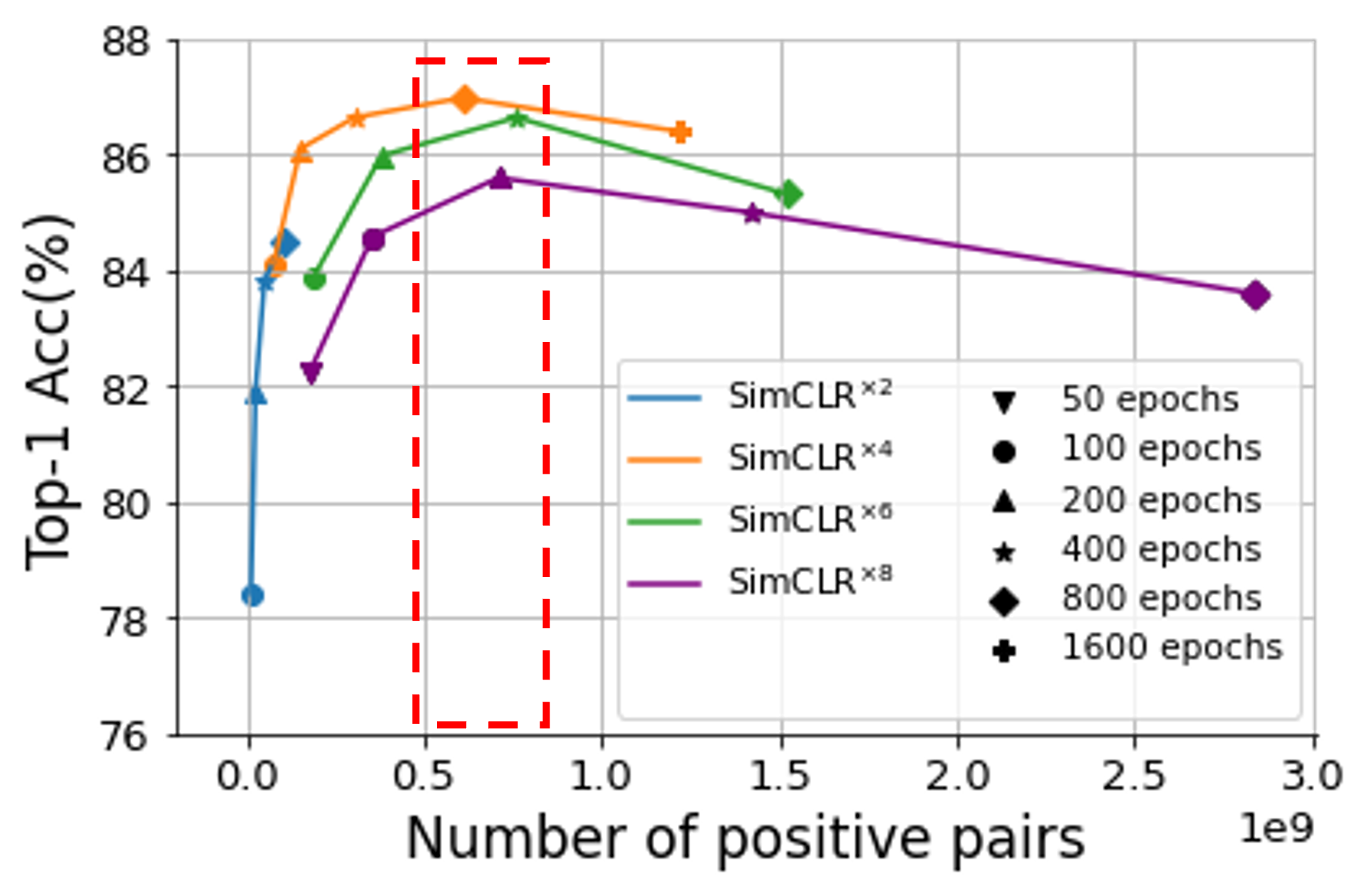}
  }
  }
\caption{\label{figure:view_performance_compare}
The effect of maximum epoch value. Because of the cosine learning rate decay~\cite{loshchilov2016sgdr}, we have evaluated the 
SimCLR$^{\times K}$ performance for a range of maximum epoch configurations. Each point in the plot corresponds to an independent evaluation. Results for ImageNet-100.}
\vspace{-0.2cm}
\end{figure}

For a more careful analysis of ECPP, we have evaluated 4 different views with 4 different training lengths, and the results are shown in Figure~\ref{figure:view_performance_compare}. In (a), it can be seen that 2-views and 4-views continue to improve their performance even at 800 epochs. But, the performance saturates earlier for 6-views and 8-views. When the same results are plotted with the number of positive pairs as the X-axis, a strong pattern can be observed. For any number of views, the best performance seems to occur in the red dotted box area in Figure~\ref{figure:view_performance_compare}(b). This indicates that it can be harmful to perform unsupervised learning for too long. Additional discussions are provided in Section~\ref{subsec:speed_discussion}.

\vspace{-0.2cm}
\section{Discussion}
\label{sec:discussion}
\vspace{-0.2cm}

\subsection{Applying ECPP to non-contrastive learning (BYOL)}
\begin{figure}[ht!]
  \centering
  \subfloat[Epoch]{
    \includegraphics[width=0.24\textwidth]{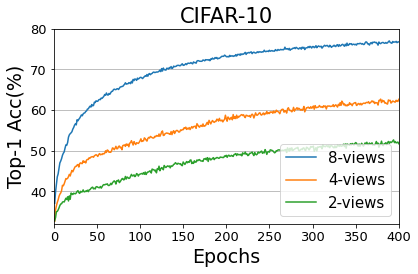}
    \includegraphics[width=0.24\textwidth]{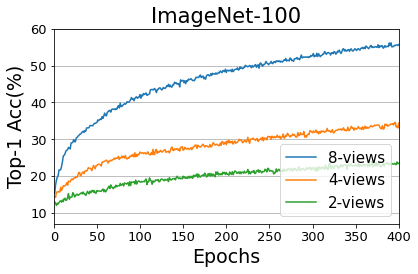}
  }
  \subfloat[Number of positive pairs]{
    \includegraphics[width=0.24\textwidth]{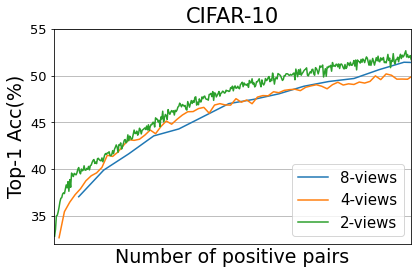}
    \includegraphics[width=0.24\textwidth]{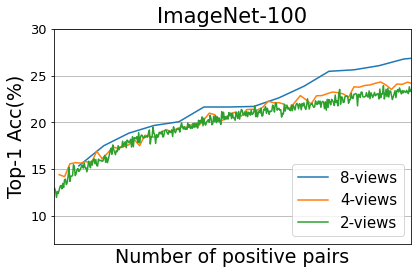}
  }
\caption{\label{figure:efficiency_curve_byol}
Linear evaluation performance of ResNet-18 with 2, 4, and 8 for BYOL$^{\times K}$. The figures are generated in the same way as in Figure~\ref{figure:efficiency_curve}.
}
\vspace{-0.2cm}
\end{figure}

\vspace{-0.1cm}
ECPP can also be combined with any non-contrastive learning algorithm as long as a positive pairing with instance augmentation is used. BYOL~\cite{grill2020bootstrap} is a very popular non-contrastive model, 
and we have combined ECPP with BYOL to demonstrate ECPP's broad applicability. In Figure~\ref{figure:efficiency_curve_byol}, it can be confirmed that BYOL$^{\times K}$ performs in a similar manner as the SimCLR$^{\times K}$ in Figure~\ref{figure:efficiency_curve}. Performance evaluation results are shown in Table~\ref{tab:rslt12}, and it can be confirmed that performance can be improved by using $K$-views.

\begin{table}[t]
\centering
\caption{Top-1\% linear evaluation of classification for BYOL$^{\times K}$ with ResNet-50. \label{tab:rslt12}}
\parbox{.49\linewidth}{
\centering
\subfloat[CIFAR-10]{
\label{tab:rslt12_a}
    \resizebox{.3\textwidth}{!}{
\begin{tabular}{@{}lrr@{}}
\toprule
Method         & Epoch    & \multicolumn{1}{l}{Performance} \\ 
\midrule
BYOL$^{\times 2}$     &         100            & 85.0                            \\
BYOL$^{\times 4}$          &          100          & 89.0                            \\
BYOL$^{\times 6}$           &            100       & 92.0                            \\
BYOL$^{\times 8}$            &              100    & 93.3                            \\ 
\bottomrule
\end{tabular}
    }
}
}
\hfill
\parbox{.49\linewidth}{
\centering
\subfloat[ImageNet-100]{
\label{tab:rslt12_b}
    \resizebox{.3\textwidth}{!}{
\begin{tabular}{@{}lrr@{}}
\toprule
Method        & Epoch & \multicolumn{1}{l}{Performance} \\ 
\midrule
BYOL$^{\times 2}$        &        100          & 75.8                            \\
BYOL$^{\times 4}$         &       100          & 81.9                            \\
BYOL$^{\times 6}$          &      100          & 81.8         \\
BYOL$^{\times 8}$           &     100          & 81.7         \\ 
\bottomrule
\end{tabular}
    }
}
}
\vspace{-0.45cm}
\end{table}

\vspace{-0.1cm}
\subsection{Performance and learning speed for a full training} 
\label{subsec:speed_discussion}
\vspace{-0.1cm}
In Section~\ref{subsec:learing_speed}, we have shown in Figure~\ref{figure:efficiency_curve} that the learning speed of ECPP is improved by $_KC_2$ times when the learning rate is a small constant and only combinatorial positive pairing is applied. Obviously, the speed and performance gain in Figure~\ref{figure:efficiency_curve} is not maintained when cosine learning rate decay and other enhancement techniques are applied to ECPP for a full training. As an evidence, the performance curves in Figure~\ref{figure:view_performance_compare}(b) do not overlap as much as in the performance curves in Figure~\ref{figure:efficiency_curve}(b). 

\begin{figure}[ht!]
\vspace{-0.25cm}
\parbox{.49\linewidth}{
\centering
  \subfloat[Epochs]{
    \includegraphics[width=0.40\textwidth]{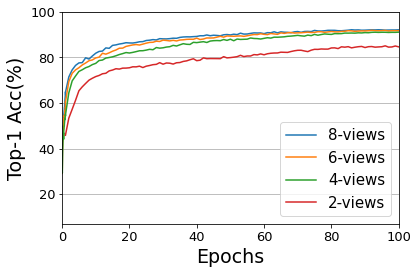}
  }
  }
\parbox{.49\linewidth}{
\centering
  \subfloat[Number of positive pairs]{
    \includegraphics[width=0.40\textwidth]{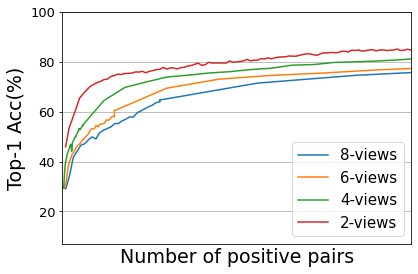}
  }
  }
\caption{Linear evaluation performance of SimCLR$^{\times K}$ for ResNet-50 on CIFAR-10 with 2, 4, 6, and 8 views. The figure (a) and (b) are generated in the similar way as in Figure~\ref{figure:efficiency_curve}. \label{fig:later_phase_effi}}
\vspace{-0.2cm}
\end{figure}

We investigated an actual full training of ECPP and the results are shown in Figure~\ref{fig:later_phase_effi}. Note that this is different from Figure~\ref{figure:view_performance_compare} because we are showing a performance curve of a single full training for each $K$. It can be confirmed from Figure~\ref{fig:later_phase_effi}(a) that larger $K$ is still helpful for a given number of epoch. The results in Figure~\ref{fig:later_phase_effi}(b), however, show that the order is reversed, i.e., smaller views perform better for a given number of positive pairs. Most likely, this reduction in efficiency is due to the correlations between the loss terms. With more views, some of the loss terms in the set $\{\mathcal{L}_{\text{}}^{V_i,V_j}\}$ inevitably become correlated and can reduce the speed gain. It is also interesting to note another limitation of ECPP shown in Figure~\ref{figure:view_performance_compare} -- the performance does not improve forever, and it peaks around a certain number of positive pairs.

\vspace{-0.2cm}
\section{Conclusion}
\label{sec:conclusion}
\vspace{-0.2cm}

In this work, we carefully study the fundamental benefits of $K$-views and propose Efficient Combinatorial Positive Pairing (ECPP), a simple add-on method that can enhance the learning speed and efficiency of contrastive learning and non-contrastive learning methods.
While contrastive and non-contrastive learning is widely adopted for a variety of unsupervised representation learning tasks, our experiments are limited to the vision tasks.
Our method can train a high-performance network with a relatively small computation. This property can be helpful to the research community.

\section*{Acknowledgements}
This work was supported by a National Research Foundation of Korea (NRF) grant funded by the Korea government (MSIT) (No. NRF-2020R1A2C2007139).

\bibliographystyle{plainnat}
\bibliography{References.bib}

\begin{thebibliography}{53}
\providecommand{\natexlab}[1]{#1}
\providecommand{\url}[1]{\texttt{#1}}
\expandafter\ifx\csname urlstyle\endcsname\relax
  \providecommand{\doi}[1]{doi: #1}\else
  \providecommand{\doi}{doi: \begingroup \urlstyle{rm}\Url}\fi

\bibitem[Alain and Bengio(2016)]{alain2016understanding}
Guillaume Alain and Yoshua Bengio.
\newblock Understanding intermediate layers using linear classifier probes.
\newblock \emph{arXiv preprint arXiv:1610.01644}, 2016.

\bibitem[Bardes et~al.(2021)Bardes, Ponce, and LeCun]{bardes2021vicreg}
Adrien Bardes, Jean Ponce, and Yann LeCun.
\newblock Vicreg: Variance-invariance-covariance regularization for self-supervised learning.
\newblock \emph{arXiv preprint arXiv:2105.04906}, 2021.

\bibitem[Caron et~al.(2020)Caron, Misra, Mairal, Goyal, Bojanowski, and Joulin]{caron2020unsupervised}
Mathilde Caron, Ishan Misra, Julien Mairal, Priya Goyal, Piotr Bojanowski, and Armand Joulin.
\newblock Unsupervised learning of visual features by contrasting cluster assignments.
\newblock \emph{Advances in Neural Information Processing Systems}, 33:\penalty0 9912--9924, 2020.

\bibitem[Chen et~al.(2020{\natexlab{a}})Chen, Kornblith, Norouzi, and Hinton]{chen2020simple}
Ting Chen, Simon Kornblith, Mohammad Norouzi, and Geoffrey Hinton.
\newblock A simple framework for contrastive learning of visual representations.
\newblock In \emph{International conference on machine learning}, pages 1597--1607. PMLR, 2020{\natexlab{a}}.

\bibitem[Chen et~al.(2020{\natexlab{b}})Chen, Kornblith, Swersky, Norouzi, and Hinton]{chen2020big}
Ting Chen, Simon Kornblith, Kevin Swersky, Mohammad Norouzi, and Geoffrey~E Hinton.
\newblock Big self-supervised models are strong semi-supervised learners.
\newblock \emph{Advances in neural information processing systems}, 33:\penalty0 22243--22255, 2020{\natexlab{b}}.

\bibitem[Chen et~al.(2021)Chen, Luo, and Li]{chen2021intriguing}
Ting Chen, Calvin Luo, and Lala Li.
\newblock Intriguing properties of contrastive losses.
\newblock \emph{Advances in Neural Information Processing Systems}, 34, 2021.

\bibitem[Chen and He(2021)]{chen2021exploring}
Xinlei Chen and Kaiming He.
\newblock Exploring simple siamese representation learning.
\newblock In \emph{Proceedings of the IEEE/CVF Conference on Computer Vision and Pattern Recognition}, pages 15750--15758, 2021.

\bibitem[Chen et~al.(2020{\natexlab{c}})Chen, Fan, Girshick, and He]{chen2020improved}
Xinlei Chen, Haoqi Fan, Ross Girshick, and Kaiming He.
\newblock Improved baselines with momentum contrastive learning.
\newblock \emph{arXiv preprint arXiv:2003.04297}, 2020{\natexlab{c}}.

\bibitem[Chuang et~al.(2020)Chuang, Robinson, Lin, Torralba, and Jegelka]{chuang2020debiased}
Ching-Yao Chuang, Joshua Robinson, Yen-Chen Lin, Antonio Torralba, and Stefanie Jegelka.
\newblock Debiased contrastive learning.
\newblock \emph{Advances in neural information processing systems}, 33:\penalty0 8765--8775, 2020.

\bibitem[Doersch et~al.(2015)Doersch, Gupta, and Efros]{doersch2015unsupervised}
Carl Doersch, Abhinav Gupta, and Alexei~A Efros.
\newblock Unsupervised visual representation learning by context prediction.
\newblock In \emph{Proceedings of the IEEE international conference on computer vision}, pages 1422--1430, 2015.

\bibitem[Gidaris et~al.(2018)Gidaris, Singh, and Komodakis]{gidaris2018unsupervised}
Spyros Gidaris, Praveer Singh, and Nikos Komodakis.
\newblock Unsupervised representation learning by predicting image rotations.
\newblock \emph{arXiv preprint arXiv:1803.07728}, 2018.

\bibitem[Goyal et~al.(2017)Goyal, Doll{\'a}r, Girshick, Noordhuis, Wesolowski, Kyrola, Tulloch, Jia, and He]{goyal2017accurate}
Priya Goyal, Piotr Doll{\'a}r, Ross Girshick, Pieter Noordhuis, Lukasz Wesolowski, Aapo Kyrola, Andrew Tulloch, Yangqing Jia, and Kaiming He.
\newblock Accurate, large minibatch sgd: Training imagenet in 1 hour.
\newblock \emph{arXiv preprint arXiv:1706.02677}, 2017.

\bibitem[Grill et~al.(2020)Grill, Strub, Altch{\'e}, Tallec, Richemond, Buchatskaya, Doersch, Avila~Pires, Guo, Gheshlaghi~Azar, et~al.]{grill2020bootstrap}
Jean-Bastien Grill, Florian Strub, Florent Altch{\'e}, Corentin Tallec, Pierre Richemond, Elena Buchatskaya, Carl Doersch, Bernardo Avila~Pires, Zhaohan Guo, Mohammad Gheshlaghi~Azar, et~al.
\newblock Bootstrap your own latent-a new approach to self-supervised learning.
\newblock \emph{Advances in Neural Information Processing Systems}, 33:\penalty0 21271--21284, 2020.

\bibitem[Harwood et~al.(2017)Harwood, Kumar~BG, Carneiro, Reid, and Drummond]{harwood2017smart}
Ben Harwood, Vijay Kumar~BG, Gustavo Carneiro, Ian Reid, and Tom Drummond.
\newblock Smart mining for deep metric learning.
\newblock In \emph{Proceedings of the IEEE International Conference on Computer Vision}, pages 2821--2829, 2017.

\bibitem[He et~al.(2016)He, Zhang, Ren, and Sun]{he2016deep}
Kaiming He, Xiangyu Zhang, Shaoqing Ren, and Jian Sun.
\newblock Deep residual learning for image recognition.
\newblock In \emph{Proceedings of the IEEE conference on computer vision and pattern recognition}, pages 770--778, 2016.

\bibitem[He et~al.(2020)He, Fan, Wu, Xie, and Girshick]{he2020momentum}
Kaiming He, Haoqi Fan, Yuxin Wu, Saining Xie, and Ross Girshick.
\newblock Momentum contrast for unsupervised visual representation learning.
\newblock In \emph{Proceedings of the IEEE/CVF conference on computer vision and pattern recognition}, pages 9729--9738, 2020.

\bibitem[Hua et~al.(2021)Hua, Wang, Xue, Ren, Wang, and Zhao]{hua2021feature}
Tianyu Hua, Wenxiao Wang, Zihui Xue, Sucheng Ren, Yue Wang, and Hang Zhao.
\newblock On feature decorrelation in self-supervised learning.
\newblock In \emph{Proceedings of the IEEE/CVF International Conference on Computer Vision}, pages 9598--9608, 2021.

\bibitem[Huynh et~al.(2022)Huynh, Kornblith, Walter, Maire, and Khademi]{huynh2022boosting}
Tri Huynh, Simon Kornblith, Matthew~R Walter, Michael Maire, and Maryam Khademi.
\newblock Boosting contrastive self-supervised learning with false negative cancellation.
\newblock In \emph{Proceedings of the IEEE/CVF Winter Conference on Applications of Computer Vision}, pages 2785--2795, 2022.

\bibitem[Iscen et~al.(2018)Iscen, Tolias, Avrithis, and Chum]{iscen2018mining}
Ahmet Iscen, Giorgos Tolias, Yannis Avrithis, and Ond{\v{r}}ej Chum.
\newblock Mining on manifolds: Metric learning without labels.
\newblock In \emph{Proceedings of the IEEE Conference on Computer Vision and Pattern Recognition}, pages 7642--7651, 2018.

\bibitem[Jacobsen et~al.(2018)Jacobsen, Smeulders, and Oyallon]{jacobsen2018revnet}
J{\"o}rn-Henrik Jacobsen, Arnold Smeulders, and Edouard Oyallon.
\newblock i-revnet: Deep invertible networks.
\newblock \emph{arXiv preprint arXiv:1802.07088}, 2018.

\bibitem[Kalantidis et~al.(2020)Kalantidis, Sariyildiz, Pion, Weinzaepfel, and Larlus]{kalantidis2020hard}
Yannis Kalantidis, Mert~Bulent Sariyildiz, Noe Pion, Philippe Weinzaepfel, and Diane Larlus.
\newblock Hard negative mixing for contrastive learning.
\newblock \emph{Advances in Neural Information Processing Systems}, 33:\penalty0 21798--21809, 2020.

\bibitem[Krizhevsky(2014)]{krizhevsky2014one}
Alex Krizhevsky.
\newblock One weird trick for parallelizing convolutional neural networks.
\newblock \emph{arXiv preprint arXiv:1404.5997}, 2014.

\bibitem[Krizhevsky et~al.(2009)Krizhevsky, Hinton, et~al.]{krizhevsky2009learning}
Alex Krizhevsky, Geoffrey Hinton, et~al.
\newblock Learning multiple layers of features from tiny images.
\newblock 2009.

\bibitem[Li et~al.(2020)Li, Zhao, Varma, Salpekar, Noordhuis, Li, Paszke, Smith, Vaughan, Damania, et~al.]{li2020pytorch}
Shen Li, Yanli Zhao, Rohan Varma, Omkar Salpekar, Pieter Noordhuis, Teng Li, Adam Paszke, Jeff Smith, Brian Vaughan, Pritam Damania, et~al.
\newblock Pytorch distributed: Experiences on accelerating data parallel training.
\newblock \emph{arXiv preprint arXiv:2006.15704}, 2020.

\bibitem[Loshchilov and Hutter(2016)]{loshchilov2016sgdr}
Ilya Loshchilov and Frank Hutter.
\newblock Sgdr: Stochastic gradient descent with warm restarts.
\newblock \emph{arXiv preprint arXiv:1608.03983}, 2016.

\bibitem[Masters and Luschi(2018)]{masters2018revisiting}
Dominic Masters and Carlo Luschi.
\newblock Revisiting small batch training for deep neural networks.
\newblock \emph{arXiv preprint arXiv:1804.07612}, 2018.

\bibitem[Micikevicius et~al.(2017)Micikevicius, Narang, Alben, Diamos, Elsen, Garcia, Ginsburg, Houston, Kuchaiev, Venkatesh, et~al.]{micikevicius2017mixed}
Paulius Micikevicius, Sharan Narang, Jonah Alben, Gregory Diamos, Erich Elsen, David Garcia, Boris Ginsburg, Michael Houston, Oleksii Kuchaiev, Ganesh Venkatesh, et~al.
\newblock Mixed precision training.
\newblock \emph{arXiv preprint arXiv:1710.03740}, 2017.

\bibitem[Misra and Maaten(2020)]{misra2020self}
Ishan Misra and Laurens van~der Maaten.
\newblock Self-supervised learning of pretext-invariant representations.
\newblock In \emph{Proceedings of the IEEE/CVF Conference on Computer Vision and Pattern Recognition}, pages 6707--6717, 2020.

\bibitem[Nguyen et~al.(2019)Nguyen, Dax, Mummadi, Ngo, Nguyen, Lou, and Brox]{nguyen2019deepusps}
Tam Nguyen, Maximilian Dax, Chaithanya~Kumar Mummadi, Nhung Ngo, Thi Hoai~Phuong Nguyen, Zhongyu Lou, and Thomas Brox.
\newblock Deepusps: Deep robust unsupervised saliency prediction via self-supervision.
\newblock \emph{Advances in Neural Information Processing Systems}, 32, 2019.

\bibitem[Noroozi and Favaro(2016)]{noroozi2016unsupervised}
Mehdi Noroozi and Paolo Favaro.
\newblock Unsupervised learning of visual representations by solving jigsaw puzzles.
\newblock In \emph{European conference on computer vision}, pages 69--84. Springer, 2016.

\bibitem[Oh~Song et~al.(2016)Oh~Song, Xiang, Jegelka, and Savarese]{oh2016deep}
Hyun Oh~Song, Yu~Xiang, Stefanie Jegelka, and Silvio Savarese.
\newblock Deep metric learning via lifted structured feature embedding.
\newblock In \emph{Proceedings of the IEEE conference on computer vision and pattern recognition}, pages 4004--4012, 2016.

\bibitem[Pathak et~al.(2016)Pathak, Krahenbuhl, Donahue, Darrell, and Efros]{pathak2016context}
Deepak Pathak, Philipp Krahenbuhl, Jeff Donahue, Trevor Darrell, and Alexei~A Efros.
\newblock Context encoders: Feature learning by inpainting.
\newblock In \emph{Proceedings of the IEEE conference on computer vision and pattern recognition}, pages 2536--2544, 2016.

\bibitem[Robinson et~al.(2020)Robinson, Chuang, Sra, and Jegelka]{robinson2020contrastive}
Joshua Robinson, Ching-Yao Chuang, Suvrit Sra, and Stefanie Jegelka.
\newblock Contrastive learning with hard negative samples.
\newblock \emph{arXiv preprint arXiv:2010.04592}, 2020.

\bibitem[Russakovsky et~al.(2015)Russakovsky, Deng, Su, Krause, Satheesh, Ma, Huang, Karpathy, Khosla, Bernstein, et~al.]{russakovsky2015imagenet}
Olga Russakovsky, Jia Deng, Hao Su, Jonathan Krause, Sanjeev Satheesh, Sean Ma, Zhiheng Huang, Andrej Karpathy, Aditya Khosla, Michael Bernstein, et~al.
\newblock Imagenet large scale visual recognition challenge.
\newblock \emph{International journal of computer vision}, 115\penalty0 (3):\penalty0 211--252, 2015.

\bibitem[Smith(2017)]{smith2017cyclical}
Leslie~N Smith.
\newblock Cyclical learning rates for training neural networks.
\newblock In \emph{2017 IEEE winter conference on applications of computer vision (WACV)}, pages 464--472. IEEE, 2017.

\bibitem[Smith and Topin(2019)]{smith2019super}
Leslie~N Smith and Nicholay Topin.
\newblock Super-convergence: Very fast training of neural networks using large learning rates.
\newblock In \emph{Artificial intelligence and machine learning for multi-domain operations applications}, volume 11006, page 1100612. International Society for Optics and Photonics, 2019.

\bibitem[Suh et~al.(2019)Suh, Han, Kim, and Lee]{suh2019stochastic}
Yumin Suh, Bohyung Han, Wonsik Kim, and Kyoung~Mu Lee.
\newblock Stochastic class-based hard example mining for deep metric learning.
\newblock In \emph{Proceedings of the IEEE/CVF Conference on Computer Vision and Pattern Recognition}, pages 7251--7259, 2019.

\bibitem[Tian et~al.(2020)Tian, Krishnan, and Isola]{tian2020contrastive}
Yonglong Tian, Dilip Krishnan, and Phillip Isola.
\newblock Contrastive multiview coding.
\newblock In \emph{European conference on computer vision}, pages 776--794. Springer, 2020.

\bibitem[Tomasev et~al.(2022)Tomasev, Bica, McWilliams, Buesing, Pascanu, Blundell, and Mitrovic]{tomasev2022pushing}
Nenad Tomasev, Ioana Bica, Brian McWilliams, Lars Buesing, Razvan Pascanu, Charles Blundell, and Jovana Mitrovic.
\newblock Pushing the limits of self-supervised resnets: Can we outperform supervised learning without labels on imagenet?
\newblock \emph{arXiv preprint arXiv:2201.05119}, 2022.

\bibitem[Touvron et~al.(2019)Touvron, Vedaldi, Douze, and J{\'e}gou]{touvron2019fixing}
Hugo Touvron, Andrea Vedaldi, Matthijs Douze, and Herv{\'e} J{\'e}gou.
\newblock Fixing the train-test resolution discrepancy.
\newblock \emph{Advances in neural information processing systems}, 32, 2019.

\bibitem[Tschannen et~al.(2019)Tschannen, Djolonga, Rubenstein, Gelly, and Lucic]{tschannen2019mutual}
Michael Tschannen, Josip Djolonga, Paul~K Rubenstein, Sylvain Gelly, and Mario Lucic.
\newblock On mutual information maximization for representation learning.
\newblock \emph{arXiv preprint arXiv:1907.13625}, 2019.

\bibitem[Wang and Liu(2021)]{wang2021understanding}
Feng Wang and Huaping Liu.
\newblock Understanding the behaviour of contrastive loss.
\newblock In \emph{Proceedings of the IEEE/CVF conference on computer vision and pattern recognition}, pages 2495--2504, 2021.

\bibitem[Wang and Isola(2020)]{wang2020understanding}
Tongzhou Wang and Phillip Isola.
\newblock Understanding contrastive representation learning through alignment and uniformity on the hypersphere.
\newblock In \emph{International Conference on Machine Learning}, pages 9929--9939. PMLR, 2020.

\bibitem[Wu et~al.(2017)Wu, Manmatha, Smola, and Krahenbuhl]{wu2017sampling}
Chao-Yuan Wu, R~Manmatha, Alexander~J Smola, and Philipp Krahenbuhl.
\newblock Sampling matters in deep embedding learning.
\newblock In \emph{Proceedings of the IEEE International Conference on Computer Vision}, pages 2840--2848, 2017.

\bibitem[Wu et~al.(2020)Wu, Mosse, Zhuang, Yamins, and Goodman]{wu2020conditional}
Mike Wu, Milan Mosse, Chengxu Zhuang, Daniel Yamins, and Noah Goodman.
\newblock Conditional negative sampling for contrastive learning of visual representations.
\newblock \emph{arXiv preprint arXiv:2010.02037}, 2020.

\bibitem[Wu and He(2018)]{wu2018group}
Yuxin Wu and Kaiming He.
\newblock Group normalization.
\newblock In \emph{Proceedings of the European conference on computer vision (ECCV)}, pages 3--19, 2018.

\bibitem[Xiao et~al.(2020)Xiao, Wang, Efros, and Darrell]{xiao2020should}
Tete Xiao, Xiaolong Wang, Alexei~A Efros, and Trevor Darrell.
\newblock What should not be contrastive in contrastive learning.
\newblock \emph{arXiv preprint arXiv:2008.05659}, 2020.

\bibitem[Xuan et~al.(2020)Xuan, Stylianou, Liu, and Pless]{xuan2020hard}
Hong Xuan, Abby Stylianou, Xiaotong Liu, and Robert Pless.
\newblock Hard negative examples are hard, but useful.
\newblock In \emph{European Conference on Computer Vision}, pages 126--142. Springer, 2020.

\bibitem[You et~al.(2017)You, Gitman, and Ginsburg]{you2017large}
Yang You, Igor Gitman, and Boris Ginsburg.
\newblock Large batch training of convolutional networks.
\newblock \emph{arXiv preprint arXiv:1708.03888}, 2017.

\bibitem[Zbontar et~al.(2021)Zbontar, Jing, Misra, LeCun, and Deny]{zbontar2021barlow}
Jure Zbontar, Li~Jing, Ishan Misra, Yann LeCun, and St{\'e}phane Deny.
\newblock Barlow twins: Self-supervised learning via redundancy reduction.
\newblock In \emph{International Conference on Machine Learning}, pages 12310--12320. PMLR, 2021.

\bibitem[Zhang et~al.(2018)Zhang, Dana, Shi, Zhang, Wang, Tyagi, and Agrawal]{zhang2018context}
Hang Zhang, Kristin Dana, Jianping Shi, Zhongyue Zhang, Xiaogang Wang, Ambrish Tyagi, and Amit Agrawal.
\newblock Context encoding for semantic segmentation.
\newblock In \emph{Proceedings of the IEEE conference on Computer Vision and Pattern Recognition}, pages 7151--7160, 2018.

\bibitem[Zheng et~al.(2021)Zheng, Chen, Yao, Yang, Li, Zhang, Zhang, Tsang, Zhou, and Zhou]{zheng2021contrastive}
Huangjie Zheng, Xu~Chen, Jiangchao Yao, Hongxia Yang, Chunyuan Li, Ya~Zhang, Hao Zhang, Ivor Tsang, Jingren Zhou, and Mingyuan Zhou.
\newblock Contrastive attraction and contrastive repulsion for representation learning.
\newblock \emph{arXiv preprint arXiv:2105.03746}, 2021.

\bibitem[Zheng et~al.(2019)Zheng, Chen, Lu, and Zhou]{zheng2019hardness}
Wenzhao Zheng, Zhaodong Chen, Jiwen Lu, and Jie Zhou.
\newblock Hardness-aware deep metric learning.
\newblock In \emph{Proceedings of the IEEE/CVF Conference on Computer Vision and Pattern Recognition}, pages 72--81, 2019.

\end{thebibliography}

\clearpage

\appendix
\begin{center}
	\LARGE{Supplementary materials for the paper \\ ``Enhancing Contrastive Learning with \\ Efficient Combinatorial Positive Pairing''}
\end{center}

\section{Implementation Details}
\label{section:implementation_details}
In this work, we use 4 $\times$ RTX 3090 GPUs as the default device for pre-training the encoding network and its evaluation.

\textbf{Encoder pre-training:} For all three datasets(CIFAR-10, ImageNet-100, and ImageNet-1K) and all view sizes,
we use base learning rate of 0.4 where the learning rate is linearly scaled with batch size ($lr$ = base learning rate $\times$ batch size / 256) and is scheduled by cosine learning rate decay with 10-epoch warm-up and without restarts~\cite{loshchilov2016sgdr,goyal2017accurate}.
We use SGD optimizer with momentum of 0.9 that is commonly adopted for ImageNet training~\cite{goyal2017accurate}.
In this work, we do not use LARS optimizer~\cite{you2017large}, because our default batch sizes is small (e.g., 64).

We use global weight decay of 0.0001 for CIFAR-10 and ImageNet-100 and 0.00002 for ImageNet-1K. Weight decay is not applied to the biases and batch normalization parameters.
We use a 2-layer MLP projection head for CIFAR-10 and a 3-layer MLP projection head for ImageNet-100 and ImageNet-1K. Each projection head's hidden-layer dimension is chosen to be the same as the encoder's representation dimension. The projection head's output dimension is always chosen to be 256.

\textbf{Linear evaluation:}
We train a single layer linear classifier on top of the frozen encoder using SGD optimizer (batch size of 256 for 100 epochs).
For CIFAR-10, we use learning rate of 0.25, momentum of 0.9, and no weight decay.
For ImageNet-100 and ImageNet-1K, we use learning rate of 0.3, momentum of 0.995, and weight decay of 0.000001.
The learning rate is scheduled by a cosine learning rate decay without warm-up and restarts.

\textbf{Temperature ($\tau$):} This work uses the temperature value of 0.2 following ~\citet{wang2021understanding} where it was shown that SimCLR can achieve a better performance with 0.2 instead of the original 0.1 that was used in SimCLR~\cite{chen2020simple}.

\textbf{Crop scale:} As the default crop scale, SimCLR~\cite{chen2020simple} and BYOL~\cite{grill2020bootstrap} use [0.08,1] and SimSiam~\cite{chen2021exploring} and MoCo-v2~\cite{chen2020improved} use [0.2,1]. We chose [0.2,1] because it performs better for the most cases shown in Table~\ref{tab:crop_scale_tau}.

\begin{table}[h!]
\caption{\label{tab:crop_scale_tau}Comparison between two different crop scale options over a variety of temperature values. Top-1\% linear evaluation results for ResNet-50, batch size of 128, and 200 epochs are shown. The results indicate that the crop scale of [0.2,1] can achieve a better performance in most of the cases and confirms that the temperature of 0.2 performs the best.}
\centering
\subfloat[CIFAR-10]{
    \resizebox{.40\textwidth}{!}{
\begin{tabular}{@{}cccccc@{}}
\toprule
$\tau$      & 0.10 & 0.15 & 0.20 & 0.25 & 0.30 \\ \midrule
(0.08,1) & 87.7 & 89.4 & 90.0 & 89.7 & 89.4 \\
(0.2,1)  & 86.9 & 89.8 & \textbf{90.4} & 90.1 & 90.0 \\ \bottomrule
\end{tabular}
}

}
\hfill
\subfloat[ImageNet-100]{
    \resizebox{.40\textwidth}{!}{
\begin{tabular}{@{}cccccc@{}}
\toprule
$\tau$      & 0.10 & 0.15 & 0.20 & 0.25 & 0.30 \\ \midrule
(0.08,1) & 76.3 & 77.6 & 78.7 & 78.3 & 77.9 \\
(0.2,1)  & 75.1 & 78.0 & \textbf{78.7} & 78.5 & 78.3 \\ \bottomrule
\end{tabular}
}

}
\end{table}

\clearpage
\textbf{Augmentation sets:} The augmentation settings are shown below. 

\begin{minted}[fontsize=\small]{python}
import torchvision.transforms as T
from PIL import ImageOps

class Solarization(object):
    def __init__(self):
        pass
    def __call__(self, img):
        return ImageOps.solarize(img)

transform_imagenet_default_large = \
T.Compose([
    T.RandomResizedCrop(224, scale=(0.2, 1.0)),
    T.RandomHorizontalFlip(),
    T.RandomApply([T.ColorJitter(0.8, 0.8, 0.8, 0.2)], p=0.8),
    T.RandomGrayscale(p=0.2),
    T.RandomApply([T.GaussianBlur(23, sigma=(0.1, 2.0))], p=0.5),
    T.RandomApply([Solarization()], p=0.1),
    T.ToTensor(),
    T.Normalize(mean=[0.485, 0.456, 0.406], std=[0.229, 0.224, 0.225]),
    ])

transform_imagenet_global_small = \
T.Compose([
    T.RandomResizedCrop(96, scale=(0.2, 1.0)),
    T.RandomHorizontalFlip(),
    T.RandomApply([T.ColorJitter(0.8, 0.8, 0.8, 0.2)], p=0.8),
    T.RandomGrayscale(p=0.2),
    T.RandomApply([T.GaussianBlur(23, sigma=(0.1, 2.0))], p=0.5),
    T.RandomApply([Solarization()], p=0.1),
    T.ToTensor(),
    T.Normalize(mean=[0.485, 0.456, 0.406], std=[0.229, 0.224, 0.225]),
    ])

transform_imagenet_global_small_crop_only = \
T.Compose([
    T.RandomResizedCrop(96, scale=(0.2, 1.0)),
    T.ToTensor(),
    T.Normalize(mean=[0.485, 0.456, 0.406], std=[0.229, 0.224, 0.225]),
    ])

# For CIFAR-10, we do not use small additional views.
transform_cifar_default = T.Compose([
    T.RandomResizedCrop(32, scale=(0.2, 1.0)),
    T.RandomHorizontalFlip(),
    T.RandomApply([T.ColorJitter(0.4, 0.4, 0.4, 0.1)], p=0.8),
    T.RandomGrayscale(p=0.2),
    T.ToTensor(),
    T.Normalize(mean=[0.4914, 0.4822, 0.4465], std=[0.2023, 0.1994, 0.2010]),
    ])

transform_cifar_global_croponly = T.Compose([
    T.RandomResizedCrop(32, scale=(0.2, 1.0)),
    T.ToTensor(),
    T.Normalize(mean=[0.4914, 0.4822, 0.4465], std=[0.2023, 0.1994, 0.2010]),
    ])
\end{minted}

\end{document}